%% file: main.tex
\documentclass{article}
\usepackage{microtype}
\usepackage{graphicx}
\usepackage{subcaption}
\usepackage{booktabs} % for professional tables
\usepackage{tabularx}
\usepackage{hyperref}
\usepackage{enumitem}

\usepackage[preprint]{icml2026}
\usepackage{amsmath}
\usepackage{amssymb}
\usepackage{mathtools}
\usepackage{amsthm}

\usepackage{caption}
\usepackage{tcolorbox}

\usepackage{array}
\usepackage{colortbl}
\usepackage{ragged2e}
\usepackage{listings}
\usepackage{tcolorbox}
\tcbuselibrary{skins}

\usepackage{pifont}
\definecolor{gradientpurple}{RGB}{107, 70, 193}  % Deep purple
\definecolor{gradientblue}{RGB}{59, 130, 246}    % Bright blue
\definecolor{lightpurple}{RGB}{139, 92, 246}     % Mid purple
\definecolor{paleblue}{RGB}{147, 197, 253}       % Light blue
\definecolor{rowalt}{HTML}{F8FAFC}

\usepackage[capitalize,noabbrev]{cleveref}

\input{commands}

\theoremstyle{plain}

\theoremstyle{definition}

\theoremstyle{remark}

\usepackage[textsize=tiny]{todonotes}

\icmltitlerunning{CUBE: A Standard for Unifying Agent Benchmarks}

\begin{document}

\twocolumn[
  \icmltitle{CUBE: A Standard for Unifying Agent Benchmarks}



  \icmlsetsymbol{equal}{*}

\begin{center}
\normalsize
\textbf{Alexandre Lacoste}\textsuperscript{1}\quad
\textbf{Nicolas Gontier}\textsuperscript{1}\quad
Oleh Shliazhko\textsuperscript{1}\quad
Aman Jaiswal\textsuperscript{1,3}\quad
Kusha Sareen\textsuperscript{1,6,7}\quad
Shailesh Nanisetty\textsuperscript{1}\quad
Joan Cabezas\quad
Manuel Del Verme\textsuperscript{2}\quad
Omar G.\ Younis\textsuperscript{2}\quad
Simone Baratta\textsuperscript{2}\quad
Matteo Avalle\textsuperscript{2}\quad
Imene Kerboua\textsuperscript{7}\quad
Xing Han L\`{u}\textsuperscript{6,7}\quad
% Felipe Vieira Frujeri\textsuperscript{4}\quad % Nvidia authors will be added back after proper internal review process
% Marc Cuevas\textsuperscript{4}\quad
% Farzan Memarian\textsuperscript{4}\quad
% Natasha Mayorga\textsuperscript{4}\quad
% Vivienne Zhang\textsuperscript{4}\quad
Elron Bandel\textsuperscript{4}\quad
Michal Shmueli-Scheuer\textsuperscript{4}\quad
Asaf Yehudai\textsuperscript{4}\quad
Leshem Choshen\textsuperscript{4}\quad
Jonathan Lebensold\textsuperscript{5}\quad
Sean Hughes\textsuperscript{1}\quad
Massimo Caccia\textsuperscript{1}\quad
Alexandre Drouin\textsuperscript{1,7}\quad
Siva Reddy\textsuperscript{6,7}\quad
Tao Yu\textsuperscript{8}\quad
Yu Su\textsuperscript{9}\quad
Graham Neubig\textsuperscript{10}\quad
Dawn Song\textsuperscript{11}\\
\vspace{6pt}
\rule{0.6\linewidth}{0.3pt}\\
\vspace{4pt}
\small
\begin{tabular}{cccc}
\textsuperscript{1}ServiceNow AI Research  & \textsuperscript{2}Silverstream.ai  & \textsuperscript{3}Dalhousie & \textsuperscript{4}IBM Research \\
\noalign{\vspace{2pt}}
\textsuperscript{5}Jetty                & \textsuperscript{6}McGill           & \textsuperscript{7}Mila    &  \\
\noalign{\vspace{2pt}}
\textsuperscript{8}HKU               & \textsuperscript{9}OSU                  & \textsuperscript{10}CMU             & \textsuperscript{11}UC Berkeley \\
\end{tabular}
\end{center}

  \icmlcorrespondingauthor{Alexandre Lacoste}{alexandre.lacoste@servicenow.com}
  \icmlcorrespondingauthor{Nicolas Gontier}{nicolas.gontier@servicenow.com}

  \icmlkeywords{Machine Learning, ICML}

  \vskip 0.3in
]



\printAffiliationsAndNotice{}  % no special notice (required even if empty)

\begin{abstract}
    The proliferation of agent benchmarks has created critical fragmentation that threatens research productivity. Each new benchmark requires substantial custom integration, creating an ``integration tax'' that limits comprehensive evaluation. \textbf{We propose CUBE (Common Unified Benchmark Environments), a universal protocol standard built on MCP and Gym that allows benchmarks to be wrapped once and used everywhere}. By separating task, benchmark, package, and registry concerns into distinct API layers, CUBE enables any compliant platform to access any compliant benchmark for evaluation, RL training, or data generation without custom integration. We call on the community to contribute to the development of this standard before platform-specific implementations deepen fragmentation as benchmark production accelerates through 2026.
\end{abstract}

% --- SECTION 1: INTRODUCTION ---
\section{Introduction}

The field of Artificial Intelligence (AI) is experiencing a remarkable surge in benchmark development for AI agents. The research community has created an impressive ecosystem of complex, interactive environments designed to test the limits of autonomous agents~\citep{yehudai2025surveyevaluationllmbasedagents, Mohammadi2025EvaluationAB}. This diversity presents tremendous opportunities: researchers can now evaluate agents across broad task distributions and leverage large-scale training on varied environments. However, realizing this potential requires solving a critical integration challenge. The effort required to incorporate these diverse benchmarks into evaluation and training pipelines has become a significant bottleneck. In practice, this limits which labs can afford to evaluate across many benchmarks, and shapes what research gets done as a result \citep{bandel2026generalagent}.

As we move toward generalist agents, the need for a general framework supporting both evaluation and post-training on diverse task distributions becomes critical. We seek agents that reason, use tools \citep{Xu2025LLMBasedAF, Shen2024LLMWT}, and navigate unseen environments. With proper tools and a consistent interface, a generalist agent should handle any benchmark \citep{bandel2026agentic}. Yet integrating new benchmarks remains time-consuming, forcing researchers to act more like systems engineers than AI scientists.

To address this integration tax, several platforms have emerged, such as NeMo Gym \citep{nemo-gym}, Harbor~\citep{Shaw_Harbor_Framework_2025}, HAL \citep{kapoor2025holisticagentleaderboardmissing}, and OpenEnv \citep{openenv2025}, among others. These platforms integrate existing benchmarks and provide tooling to author new environments, but each proposes its own environment interface. As a result, maintainers must build connectors to move benchmarks across them, a growing burden as benchmark production accelerates. Recognizing this, the community has also begun proposing open benchmark interface standards: the Agentified Agent Assessment (AAA) paradigm in AgentBeats \citep{agentbeats2025} and CUBE (this paper) are concurrent efforts sharing this motivation, though differing in scope and design choices, as discussed in Section~\ref{sec:related}.

Competition between platforms is a valuable market force and drives innovation. They should compete on features, usability, scalability, and other value metrics. Yet without a common benchmark interface, competition skews toward the size of their integration catalog instead.

\definecolor{royalblue}{RGB}{65, 105, 225}

\begin{tcolorbox}[
  enhanced,
  colback=royalblue!8,
  colframe=royalblue!90,
  title={\textbf{The Core Position}},
  fonttitle=\bfseries\large,
  coltitle=white,
  arc=1mm,
]
\textbf{The community needs a standard that allows practitioners to wrap agentic benchmarks once and have it work everywhere, for evaluation and for training, at scale.}
\end{tcolorbox}

Any platform that implements the standard would instantly gain access to every compliant benchmark. This would eliminate redundant integration work and let us focus on understanding and improving agent behavior. By moving toward a universal standard, we can turn the current landscape of isolated silos into a thriving, interoperable ecosystem.

The importance of multi-benchmarking cannot be overstated. There are currently over 300 agentic benchmarks available, many of which are highly innovative but remain largely unknown because they are too difficult to set up. By simplifying and scaling cross-benchmark evaluation, we can gain a much deeper understanding of agent capabilities across a vast distribution of tasks. Given the rapid rise of coding agents \citep{yang2024swebenchverified, openhands2024, jimenez2023swebench, Deng2025SWEBenchPC} and the increasing interest in post-training and RL on diverse tasks \citep{Khatri2025TheAO, Park2025MAPoRLMP}, we forecast this number to double by the end of 2026. Without a standard, the field is destined for unmanageable fragmentation as benchmark production outstrips integration capacity.

% --- SECTION 2: THE CURRENT SITUATION ---
\section{The Current Landscape and its Challenges}

% Caption setup
\captionsetup[table]{position=top, font=small, labelfont=bf}

\begin{table*}[t]
\centering
\caption{A comparative analysis of the infrastructure and operational requirements across current agentic benchmarks. The lack of uniformity in how these environments are hosted, controlled, and reset creates a significant overhead for cross-benchmark evaluation.}
\label{tab:benchmark_comparison}
\scriptsize  % Even smaller font
\renewcommand{\arraystretch}{1.25}
\rowcolors{2}{rowalt}{white}
\resizebox{\linewidth}{!}{
\begin{tabular}{>{\raggedright\hangindent=0.5em}p{1.5cm} 
                >{\raggedright\hangindent=0.5em}p{3.6cm} 
                >{\raggedright\hangindent=0.5em}p{3.4cm} 
                >{\raggedright\hangindent=0.5em}p{3.4cm} 
                >{\raggedright\hangindent=0.5em\arraybackslash}p{3.2cm}}
\toprule
\textbf{Feature} & \textsc{WebArena} \newline \citep{zhou2024webarena} & \textsc{SWE-bench} \newline\citep{jimenez2023swebench} & \textsc{osworld} \newline\citep{xie2024osworld} & \textsc{GAIA} \newline \citep{mialon2023gaia} \\
\midrule
\textbf{Environment Type} & \textbf{Simulated Web}: A private micro-internet (GitLab, Reddit clone, etc.). & \textbf{Coding Workspace}: A specific software repository and dev toolchain. & \textbf{Desktop OS}: A full graphical operating system (Ubuntu/Windows). & \textbf{The Real World}: Live internet access and local document sets. \\
\addlinespace[0.5em]
\textbf{Hosting Format} & \textbf{Benchmark-level VM}: One shared world per task run. High AWS costs. & \textbf{Task-level Container}: Lightweight, ephemeral Docker images per task. & \textbf{Benchmark-level VM}: OS VM with Task-level Snapshots. & \textbf{Static Files}: Task files on HF Hub; users provide tool implementations. \\
\addlinespace[0.5em]
\textbf{Action Space (Tooling)} & \textbf{Flexible/Provided}: Playwright basics provided, but usually swapped for custom abstractions (e.g., BrowserGym). & \textbf{Fixed Shell}: Standard terminal-based actions (Bash/Git). & \textbf{Native GUI}: Hard-coded mouse/keyboard coordinate actions. & \textbf{None (BYOT)}: Users must build tools (Search, PDF reader) from scratch. \\

\addlinespace[0.5em]

\textbf{Integration Effort} & \textbf{High}: Requires solving rigid networking (port mapping) and building a perception bridge to the HTML. & \textbf{Moderate}: Standard Docker orchestration; agents speak Terminal natively. & \textbf{Moderate}: Authors provide the VM/API, but vision-to-action grounding is a high hurdle. & \textbf{High}: The researcher must provide a broad range of tools. \\
\addlinespace[0.5em]
\textbf{Scalability Bottleneck} & \textbf{State Reset Latency:} VMs require high resources. Benchmark recommends sequential evaluation of all tasks.  
& \textbf{Disk I/O Churn}: Constant building/pulling of unique task images saturates the filesystem. & \textbf{RAM \& Snapshots}: 20\,GB+ memory per agent and heavy disk I/O for state resets. & \textbf{API \& Rate Limits}: Capped by external search quotas and LLM token costs. \\

\bottomrule
\end{tabular}
}
\end{table*}

To understand why a standard is necessary, consider the engineer bridging a benchmark and a generalist agent, translating raw implementations into agent-compatible interfaces through technical hurdles rarely discussed in papers.

Unlike static datasets, agentic benchmarks require live, interactive environments \citep{appworld2024, merrill2026terminalbenchbenchmarkingagentshard} with diverse infrastructure needs spanning web navigation \citep{drouin2024workarena, zhou2024webarena, Koh2024VisualWebArenaEM}, software engineering \citep{jimenez2023swebench, Li2024DevBenchAC}, operating systems \citep{xie2024osworld, bonatti2025windows}, and mobile devices \citep{rawles2025androidworld, chen2025spabench}. Some require shared servers common to all tasks, preventing simple per-task deployment. To make it concrete, Table~\ref{tab:benchmark_comparison} summarizes the shape and challenges of 4 popular benchmarks.

Furthermore, deployment on high-performance computing platforms often runs into port configuration conflicts. Default ports are frequently blocked, requiring custom reconfiguration for every single deployment. Resource scaling also varies wildly, from lightweight scripts to heavy simulation environments that demand massive amounts of RAM and disk I/O. The result is an environment where reproducibility is hampered by the sheer complexity of the underlying stack.

The diversity of these requirements creates what we call the Integration Tax. When a researcher wants to evaluate an agent on five different benchmarks, they often have to write five unique ``drivers'' or ``wrappers.'' If they decide to switch from one evaluation framework to another, they must often start this work from scratch. This N-to-M mapping of agents to benchmarks is a massive waste of human capital.

% --- SECTION 3: THE CUBE STANDARD PROPOSAL ---
\section{The CUBE Standard Proposal}
\label{sec:cube_proposal}

We propose CUBE (Common Unified Benchmark Environments), a protocol standard designed to unify the ML Community by establishing a universal interface between benchmarks and evaluation frameworks.\footnote{Reference implementation: \href{https://github.com/The-AI-Alliance/cube-standard}{\texttt{github.com/The-AI-Alliance/cube-standard}} \citep{cube-standard-repo}} The core insight is simple: if we define a consistent API contract, any CUBE-compliant benchmark becomes immediately usable by any CUBE-compliant platform. Whether for evaluation or post-training, with minimal custom integration work. Most importantly, we communicate this early to allow the community to steer the standard before it reaches a more rigid state.

\begin{figure}[htbp]
  \centering
  \includegraphics[width=\columnwidth]{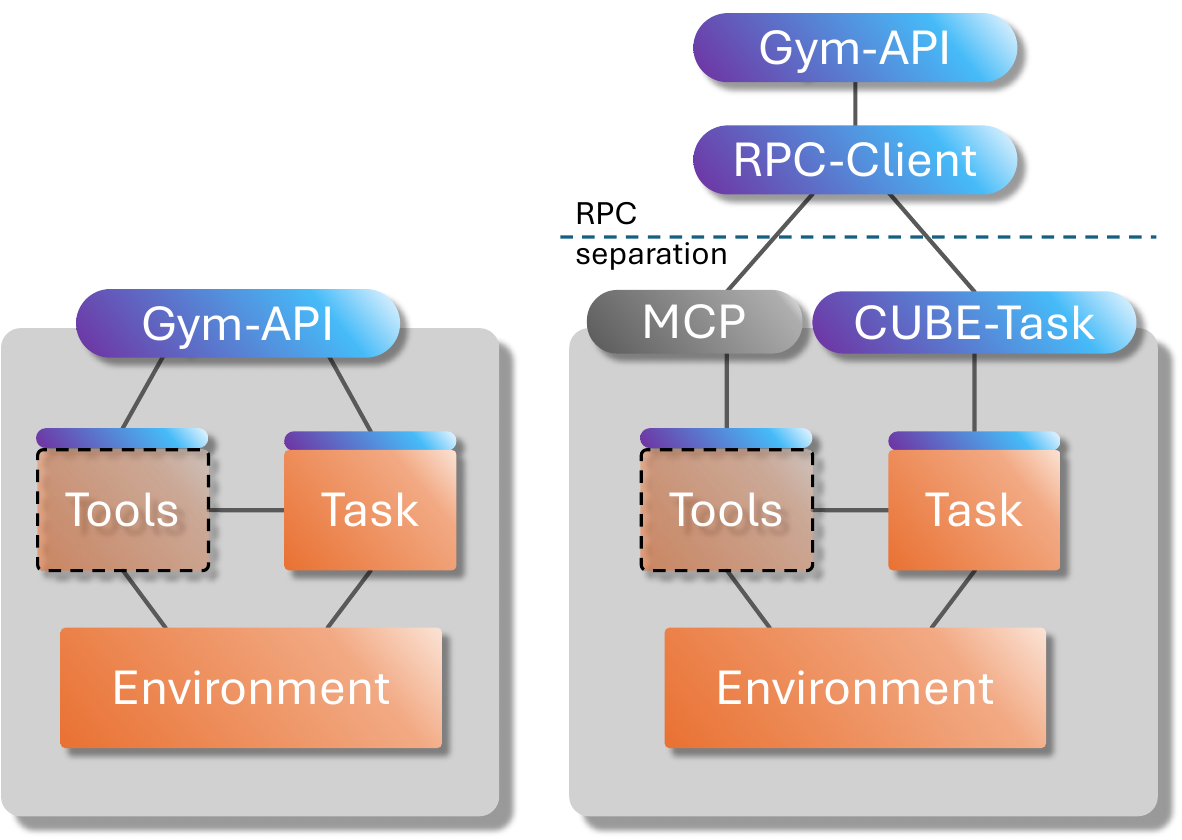}
  \caption{Task-level diagram for CUBE's API. \textbf{Left:} Separation between tasks and tools and the possibility to reconfigure the tools. Only the Gym-API is exposed. \textbf{Right:} By implementing base classes, CUBE's implementation can automatically expose an RPC layer, including the well-known MCP API. On the client side, the Gym-API is exposed.}
  \label{fig:cube_diagram}
\end{figure}

For generality and clarity, we present CUBE's interface using RPC (Remote Procedure Call) notation, but the standard supports both a Python and RPC interface (see Sec.~\ref{sec:python_first} and Fig.~\ref{fig:cube_diagram}). The RPC layer enables process isolation and cross-language communication, while the direct Python interface reduces cross-process serialization overhead and may lead to necessary speedup for some post-training applications.

\textbf{Four-layer Schema:} To achieve a standard where a generalist agent could interact with a new benchmark with minimal to no human involvement, we are mindful of 4 different levels of interaction:
\begin{enumerate}[nosep,leftmargin=*,topsep=0pt,partopsep=0pt,parsep=0pt,itemsep=0pt]
    \item \textbf{Task level:} We define how agents interact with individual task instances, how they observe state, execute actions, and receive feedback (Sec.~\ref{sec:task_level}).
    \item \textbf{Benchmark level:} We specify how evaluation harnesses discover available tasks and spawn new instances (Sec.~\ref{sec:benchmark_level}).
    \item \textbf{Package level:} We standardize installation and parallelization across compute infrastructures (Sec.~\ref{sec:package_level}).
    \item \textbf{Registry level:} We provide a centralized metadata catalog for discovery and filtering (Sec.~\ref{sec:registry})
\end{enumerate}
Each layer can be accessed either via direct Python calls for same-process execution or via RPC for distributed and cross-platform scenarios.

This separation of concerns is deliberate. A benchmark author implements their environment once in Python, exposing the required methods as a standard class. The CUBE framework automatically provides the RPC wrapper, meaning the benchmark becomes immediately usable both locally and remotely without additional work. A platform developer can point their harness at any CUBE benchmark and immediately begin evaluation, choosing between local Python instantiation for performance or RPC connection for flexibility. A researcher can filter and install benchmarks based on their available compute resources, without reading documentation or reverse-engineering setup scripts.

\subsection{Task-Level Interface}
\label{sec:task_level}

The task-level interface defines how agents interact with individual task instances. At its core, an agent needs to observe the environment's state, execute actions, and receive feedback on its progress. The Gym interface \citep{towers2024gymnasium, Brockman2016OpenAIG} established this pattern for RL, but modern agent benchmarks introduce new requirements that demand extensions to the traditional blocking \texttt{step} function.

\textbf{The Async Problem:} Consider an agent navigating a research task that requires web search. When the agent calls a search API, waiting 3 seconds for results while blocking all other operations is inefficient. The agent should be able to plan its next move, process partial results, or manage multiple concurrent tool calls. Benchmarks like ARE \citep{froger2025are} and GAIA-2 \citep{meta2026gaia2} explicitly require this asynchronous capability, where agents coordinate multiple long-running operations without blocking. The traditional Gym \texttt{step} function, designed for synchronous state transitions, cannot support these patterns.

\textbf{MCP + Gym Fusion:} CUBE addresses this by building on the Model Context Protocol (MCP), which already defines a non-blocking \texttt{tools/call} API for asynchronous action execution. MCP also handles automatic action space discovery through \texttt{tools/list}\footnote{Gym also provides \texttt{action\_space}, but MCP's format is more aligned with LLM expectations}, eliminating the need for agents to know tool signatures in advance. We augment MCP with Gym-style evaluation semantics: \texttt{cube/evaluate} returns reward and termination status, \texttt{cube/reset} reinitializes tasks with optional seeding, and \texttt{cube/close} handles cleanup. This fusion creates a superset where both established APIs can be recovered, and a new async-Gym API supports non-blocking interaction patterns.

\textbf{Tool Configuration:} Many benchmarks are designed agnostically to the specific tools used for task completion. For example, WebArena \citep{zhou2024webarena} defines the environment, task descriptions, and evaluation functions, but leaves browser interaction mechanisms to the agent designer. This design choice enables comparative evaluation of different browser automation tools. However, without specified tools, benchmarks cannot expose Gym or MCP interfaces. CUBE resolves this by requiring benchmark authors to wrap their environments with default tools, enabling immediate usability. To preserve tool variability for research, benchmarks accept a \texttt{tool\_config} parameter at initialization (see Section~\ref{sec:package_level}), allowing researchers to substitute alternative tool implementations without modifying the benchmark code.

Benchmarks provide domain-appropriate tools matched to their environment types. Browser-based benchmarks ship with web automation tools, coding benchmarks provide shell access, and GUI benchmarks offer mouse and keyboard control. Researchers can reconfigure these tools at benchmark initialization through the \texttt{tool\_config} parameter (see Section~\ref{sec:package_level}), enabling experiments with different tool implementations without modifying the benchmark itself.

\textbf{Privileged Information for Evaluation:} Beyond basic task execution, CUBE supports scalable evaluation infrastructure through privileged information. As task repositories expand, manual analysis of agent behavior becomes infeasible, leading practitioners to rely on automated judge-based evaluation for large-scale failure analysis. However, LLM-based judges suffer from well-documented limitations \citep{lu2025agentrewardbench} and frequently misidentify failure root causes. To enhance judge accuracy, we standardize the communication of privileged information at both the task and step levels through the \texttt{info} field. This information may include the evaluation function's source code, ground-truth answers, or concise summaries of the environment's internal state. Benchmark designers curate this optional field to facilitate more accurate failure diagnosis. Beyond evaluation, privileged information enables privileged policy distillation during training, where a student policy learns from a teacher with access to additional context, and aids in identifying benchmarks with erroneous evaluation logic or ambiguous task specifications.

Table \ref{tab:task-api} details the complete API. The MCP methods handle action execution and tool discovery. The CUBE-specific methods extend this with evaluation semantics and privileged information access.

% Colors
\definecolor{mcpgray}{HTML}{6B7280}
\definecolor{cubepurple}{HTML}{7C3AED}

% Column types
\newcolumntype{L}[1]{>{\RaggedRight\arraybackslash}p{#1}}
\newcolumntype{C}[1]{>{\centering\arraybackslash}p{#1}}

\renewcommand{\arraystretch}{1.4}

\begin{table*}[t]
\centering
\caption{Task-Level API. Methods available as both Python class methods (e.g., \texttt{task.reset()}) and RPC endpoints (e.g., \texttt{cube/reset}). The first part is the well-known MCP protocol. The second part adds observation, step, and evaluation functions aligned with the Gym API.}
\footnotesize
\begin{tabular}{L{2.4cm} C{0.8cm} C{1.1cm} L{2.4cm} L{4cm} L{3cm}}

\toprule
\textbf{Method} & \!\!\!\!\textbf{Namespace} & \textbf{For} & \textbf{Arguments} & \textbf{Returns} & \textbf{Description} \\
\midrule

\rowcolor{rowalt}
{\scriptsize\texttt{tools/list}} & \textcolor{mcpgray}{\textsc{mcp}} & Agent & \textit{none} & \texttt{Tool[]} & List actions \\

{\scriptsize\texttt{tools/call}} & \textcolor{mcpgray}{\textsc{mcp}} & Agent & \texttt{\{name, args\}} & \texttt{\{content[], isError\}} & Execute action \\

\rowcolor{rowalt}
{\scriptsize\texttt{resources/list}} & \textcolor{mcpgray}{\textsc{mcp}} & Agent & \textit{none} & \texttt{Resource[]} & List resources \\

{\scriptsize\texttt{resources/read}} & \textcolor{mcpgray}{\textsc{mcp}} & Agent & \texttt{\{uri\}} & \texttt{\{contents[]\}} & Read obs/task \\

\midrule

\rowcolor{rowalt}
{\scriptsize\texttt{cube/evaluate}} & \textcolor{cubepurple}{\textsc{cube}} & Harness & \textit{none} & {\scriptsize\texttt{\{obs, reward, terminated, truncated, info\}}} & Eval state \\

{\scriptsize\texttt{cube/reset}} & \textcolor{cubepurple}{\textsc{cube}} & Harness & \texttt{\{seed?\}} & \texttt{\{obs, info\}} & Reset task \\  \rowcolor{rowalt} {\scriptsize\texttt{cube/step}} & \textcolor{cubepurple}{\textsc{cube}} & Agent & \texttt{\{action\}} & {\scriptsize\texttt{\{obs, reward, terminated, truncated, info\}}} & Execute action and evaluate \\

\rowcolor{rowalt}
{\scriptsize\texttt{cube/close}} & \textcolor{cubepurple}{\textsc{cube}} & Harness & \textit{none} & \texttt{void} & Cleanup \\

{\scriptsize\texttt{cube/privileged\_info}} & \textcolor{cubepurple}{\textsc{cube}} & Harness & \textit{none} & \texttt{String} & Privileged context for judge-based evaluation \\

\bottomrule

\end{tabular}
\label{tab:task-api}
\end{table*}
% ------------

\subsection{Benchmark-Level Interface}
\label{sec:benchmark_level}

\begin{table*}[t]
\centering
\caption{Benchmark-Level API. Methods available as both Python class methods and RPC endpoints for discovering benchmarks, listing tasks, and orchestrating task instances. Each spawned task exposes its own Task-Level API endpoint.}
\footnotesize
\begin{tabular}{L{2.2cm} C{0.8cm} C{1.1cm} L{2.4cm} L{4cm} L{3cm}}
\toprule
\textbf{Method} & \!\!\!\!\!\textbf{Namespace} & \textbf{For} & \textbf{Arguments} & \textbf{Returns} & \textbf{Description} \\
\midrule
\rowcolor{rowalt}
{\scriptsize\texttt{cube/info}} & \textcolor{cubepurple}{\textsc{cube}} & Harness & \textit{none} & \texttt{BenchmarkInfo} & Benchmark metadata \\
{\scriptsize\texttt{cube/tasks}} & \textcolor{cubepurple}{\textsc{cube}} & Harness & \texttt{\{filter?\}} & \texttt{TaskList} & List available tasks \\
\rowcolor{rowalt}
{\scriptsize\texttt{cube/spawn}} & \textcolor{cubepurple}{\textsc{cube}} & Harness & \texttt{\{task\_id, seed?\}} & \texttt{url} & Start task, return endpoint \\
{\scriptsize\texttt{cube/status}} & \textcolor{cubepurple}{\textsc{cube}} & Harness & \textit{none} & \texttt{TaskStatus[]} & Health of running tasks \\
\rowcolor{rowalt}
{\scriptsize\texttt{cube/shutdown}} & \textcolor{cubepurple}{\textsc{cube}} & Harness & \texttt{\{session\_id?\}} & \texttt{void} & Cleanup all or specific \\
\bottomrule
\end{tabular}
\label{tab:benchmark-api}
\end{table*}

The benchmark-level interface (also available as both Python methods and RPC endpoints) manages shared infrastructure and orchestration. 
While the task-level interface handles individual agent-environment interactions, many benchmarks require shared infrastructure that spans multiple tasks. WebArena \citep{ zhou2024webarena}, for instance, deploys a persistent set of web services (GitLab, e-commerce sites, forums) that form a coherent ``micro-internet''. OSWorld \citep{xie2024osworld} maintains a full desktop operating system with pre-installed applications. These benchmark-level resources are expensive to initialize and are designed to be reused across many task instances.

The benchmark-level interface exists to manage this shared infrastructure and to provide discovery and orchestration capabilities. Table \ref{tab:benchmark-api} specifies the required methods. The \texttt{cube/info} endpoint returns metadata about the benchmark, including its name, version, and resource requirements. The \texttt{cube/tasks} method lists available tasks, supporting pagination and filtering for benchmarks with large task sets.

Task instantiation occurs through \texttt{cube/spawn}, which accepts a task identifier and an optional random seed. The seed parameter is required for benchmarks with stochastic task generation or variable initial states. For instance, a benchmark might generate synthetic websites with randomized layouts, requiring seed control for reproducibility. Upon spawning, the benchmark returns a URL endpoint for the new task instance. This endpoint exposes the task-level API described in Section~\ref{sec:task_level}. The separation between benchmark and task endpoints enables efficient resource sharing. A single benchmark server can manage dozens of concurrent task instances, each with its own isolated state, while sharing the underlying infrastructure. The task-level API naturally supports asynchronous execution patterns and multi-agent scenarios: multiple agents can interact with the same task instance or separate instances simultaneously, with the benchmark coordinating state updates and turn-taking as needed.

The \texttt{cube/status} method provides health monitoring for active tasks, reporting resource usage and connection status. Finally, \texttt{cube/shutdown} handles cleanup, accepting an optional session identifier to terminate specific tasks or, if omitted, shutting down all active instances.

\subsection{Package-Level Standard}
\label{sec:package_level}

The main responsibility of the package-level standard is to expose a hook for starting the RPC server (if requested) and initializing the common resources. After package installation, users or harnesses can use the following Python API:

% The package-level interface addresses benchmark deployment across diverse compute environments. Rather than prescribing deployment strategies, CUBE uses a capability-driven approach: benchmarks declare their requirements and provide standard entry points, while harnesses make infrastructure and parallelization decisions.

% \subsubsection{Capability Declaration}

% Every CUBE package exposes metadata describing its deployment characteristics. The \texttt{parallelization} field indicates whether tasks require shared infrastructure (\texttt{shared-server}), can run independently (\texttt{task-independent}), or support both (\texttt{hybrid}). Additional fields specify resource requirements (RAM, GPU, disk), network needs (internet access, port count), and supported deployment modes. This information enables harnesses to filter compatible benchmarks and plan deployment strategies.

% \subsubsection{Standard Interface}

% CUBE packages provide a Python interface for instantiation:

\begin{tcolorbox}[
  colback=blue!3,
  colframe=blue!25,
  arc=2mm,
  boxrule=0.5pt,
  left=3mm,
  right=3mm,
  top=2mm,
  bottom=2mm,
  boxsep=0pt
]
\begin{lstlisting}
import my_cube
benchmark = my_cube.Benchmark()
benchmark.start(available_ports, tool_config)
\end{lstlisting}
\end{tcolorbox}

% \begin{tcolorbox}[
%   colback=blue!3,
%   colframe=blue!25,
%   arc=2mm,
%   boxrule=0.5pt,
%   left=3mm,
%   right=3mm,
%   top=2mm,
%   bottom=2mm,
%   boxsep=0pt
% ]
% \begin{minted}[
%   style=pastie,
%   fontsize=\scriptsize,
%   breaklines,
%   bgcolor={},
%   baselinestretch=1.15
% ]{python}
% import my_cube
% bench = my_cube.Benchmark()
% task_config = bench.get_debug_task_config()
% debug_agent = bench.get_debug_agent(task_config)
% \end{minted}
% \end{tcolorbox}
A corresponding command-line interface is also exposed for non-Python access.

\textbf{Separating What from How:} CUBE separates \emph{what} a benchmark requires from \emph{how} those resources are provisioned. Benchmark authors declare resource requirements through typed configuration objects (\texttt{VMConfig}, \texttt{ContainerConfig}), while harness operators supply a matching backend that handles provisioning. Pluggable backends cover the full range of compute environments, from local development to cloud containers, major cloud VM providers, and SLURM-based HPC clusters. Switching from local to cloud requires changing a single backend configuration object, with no benchmark code changes. CUBE also distinguishes \emph{benchmark-level} shared resources (e.g., the persistent web server that WebArena shares across all task instances) from \emph{task-level} resources (e.g., a per-task container). Benchmark authors initialize shared resources once in \texttt{Benchmark.setup()}, making them available through a \texttt{RuntimeContext} passed to each task.

\par\medskip\noindent\textbf{Debug Tasks and Debug Agent:} To ensure correctness and enable continuous integration, every CUBE package must expose two additional elements at the Python level:
\texttt{get\_debug\_task\_configs()}, which returns a small set of representative task configurations with known correct behavior, and \texttt{make\_debug\_agent(task\_id)}, which returns a scripted agent guaranteed to solve a given debug task. These primitives allow any consumer of the benchmark, or the CUBE compliance suite itself, to run a full episode end-to-end and assert that the reward reaches 1.0, without requiring a live language model. This makes CUBE benchmarks testable in standard CI pipelines.\par\medskip\noindent\textbf{Stress Testing and Compliance:} Beyond basic correctness, CUBE defines a stress test suite that validates benchmark behavior under parallel load and resource constraints. This includes verifying that task resets are idempotent, that concurrent task instances remain isolated, and that resource usage stays within declared bounds. A benchmark that passes the stress suite earns a compliance badge visible in the registry, giving platform developers confidence before large-scale integration.

\subsection{CUBE Registry}
\label{sec:registry}

The CUBE Registry serves as a centralized discovery mechanism for available benchmarks. Without a registry, researchers must rely on word-of-mouth, social media announcements, or manual literature searches to find relevant evaluation environments. This creates a significant barrier for newer benchmarks, which may remain unknown despite their technical merit simply because they lack visibility in the community.

Table \ref{tab:registry-fields} specifies the metadata required for each registered benchmark. Beyond basic identification fields like name and version, the registry captures critical operational information that enables automated filtering. The \texttt{runtime} field specifies the deployment model (Docker, Apptainer, VM, or live internet access), allowing researchers to immediately exclude benchmarks incompatible with their infrastructure. The \texttt{hardware} object details resource requirements, preventing researchers from attempting to run memory-intensive benchmarks on constrained systems.

The registry also addresses legal and compliance concerns that often block benchmark adoption. The \texttt{package\_license} and \texttt{benchmark\_license} fields distinguish between the wrapper code license and the underlying task data license, as these often differ. The \texttt{content\_notice} field warns about special considerations, such as benchmarks containing cloned websites or copyrighted materials. 

Crucially, the registry does not host benchmark code or data. It simply indexes metadata and points to standard distribution platforms like PyPI via the \texttt{package} field. This design keeps the registry lightweight and ensures that benchmark authors retain full control over their distributions. To ensure quality, once a benchmark is registered, it triggers a GitHub job to verify compliance.

This automated discovery mechanism democratizes benchmark visibility. A graduate student publishing a novel benchmark can register it once and immediately make it discoverable to the entire community, without requiring social media threads, blog posts, or conference presentations to gain adoption. This enables broader evaluation across diverse benchmark suites \citep{Liu2023AgentBenchEL, Ma2024AgentBoardAA, Wang2024GTAAB} and tool-use scenarios \citep{Wang2025MCPBenchBT, Lei2025MCPVerseAE, Luo2025MCPUniverseBL}.
\renewcommand{\arraystretch}{1.4}
\begin{table*}[t]
\centering
\caption{Registry Fields. Each registered benchmark exposes this metadata for discovery, installation, and compliance verification.}
\scriptsize
\rowcolors{2}{rowalt}{white}
\begin{tabular}{L{2.8cm} L{2.2cm} L{8.5cm}}
\toprule
\textbf{Field} & \textbf{Type} & \textbf{Description} \\
\midrule
\texttt{id} & \texttt{string} & Unique identifier (e.g., \texttt{webarena-verified}) \\
\texttt{name} & \texttt{string} & Human-readable name \\
\texttt{version} & \texttt{string} & Semantic version (e.g., \texttt{1.2.0}) \\
\texttt{authors} & \texttt{string[]} & Package authors \\
\texttt{paper} & \texttt{string?} & Related paper URL (if any) \\
\texttt{package} & \texttt{string} & PyPI package name for \texttt{pip install} \\
\texttt{benchmark\_license} & \texttt{string} & Benchmark data/tasks license (e.g., \texttt{CC-BY-NC-4.0}) \\
\texttt{content\_notice} & \texttt{string?} & Copyright warning (e.g., \texttt{"Contains cloned websites"}) \\
\texttt{compliance} & \texttt{string[]} & Compliance badges (e.g., \texttt{["no-docker-root", "task-isolated"]}) \\
\texttt{runtime} & \texttt{enum} & \texttt{docker} \textbar{} \texttt{apptainer} \textbar{} \texttt{vm} \textbar{} \texttt{docker-root} \textbar{} \texttt{docker-in-docker} \textbar{} \texttt{live} \\
\texttt{hardware} & \texttt{object} & \texttt{\{ram\_gb, gpu, disk\_gb\}} \\
\texttt{task\_count} & \texttt{int} & Number of tasks in benchmark \\
\bottomrule
\end{tabular}
\label{tab:registry-fields}
\end{table*}

\subsection{Python-First Design with RPC Fallback}
\label{sec:python_first}

CUBE supports both local (same-process) and remote (RPC) execution through a unified interface. Benchmark authors implement Python classes; CUBE auto-generates RPC servers exposing identical methods over HTTP. Switching modes requires only connection changes.

Local execution eliminates serialization, reducing latency for high-frequency RL loops and enabling unified debugging. Remote execution handles: (1) non-containerizable production benchmarks, (2) cross-platform scenarios, (3) multi-language ecosystems, and (4) fault isolation.

\subsection{Adoption Strategy}
\label{sec:adoption}

CUBE faces a classic two-sided adoption challenge: platforms will hesitate to implement the standard without a critical mass of compliant benchmarks, and benchmark authors will hesitate to wrap their environments without platform demand. We break this deadlock by recruiting an initial consortium of early platform supporters who are committed to implementing reference connectors, while simultaneously wrapping a high-value corpus that provides those platforms with immediate utility. 

\paragraph{Call for Collaboration}
The authorship of this proposal spans major technology companies, academic laboratories, and startups, organizations that independently converged on the Integration Tax as a bottleneck to their own research. This convergence is itself evidence that the problem demands a coordinated response rather than competing proprietary solutions.

\paragraph{Initial implementation}
We propose to deliver a reference architecture of the CUBE standard as a starting point for community feedback, alongside an initial corpus of wrapped benchmarks spanning web navigation, software engineering, and desktop environments. By building reference connectors for platforms such as NVIDIA NeMo Gym and OpenEnv, we ensure that this corpus has immediate consumability in existing training and evaluation infrastructure. This dual initiative—benchmarks and connectors—is designed to demonstrate the standard's value and enable immediate collaboration with the broader community.

\paragraph{Direct Benchmark Creator Outreach}
We will directly engage authors of recently published benchmarks, offering integration support and highlighting the viability benefits of registry inclusion. For benchmark authors, CUBE solves the discovery problem: wrap once, and gain immediate access to the ecosystems of every training and evolution platform with CUBE-support. We will provide integration templates and registry submission assistance to minimize barriers, targeting critical mass by the end of 2026.

\section{Related Work}
\label{sec:related}

\begin{table*}[!t]
\centering
\caption{Comparison of CUBE with existing agent benchmark frameworks, based on features and documentation at the time of writing.}
\label{tab:comparison}
\scriptsize
\rowcolors{2}{rowalt}{white}
\resizebox{\linewidth}{!}{
\begin{tabular}{@{}>{\raggedright\arraybackslash}p{1.8cm}>{\raggedright\arraybackslash}p{2.7cm}>{\raggedright\arraybackslash}p{2.7cm}>{\raggedright\arraybackslash}p{2.7cm}>{\raggedright\arraybackslash}p{2.7cm}>{\raggedright\arraybackslash}p{2.7cm}@{}}
\toprule
\textbf{Feature} & \textbf{CUBE} & \textbf{NeMo Gym} & \textbf{AgentBeats} & \textbf{OpenEnv} & \textbf{Harbor} \\
\midrule

Primary Focus
& Protocol standard for wrapping benchmarks once, usable for evaluation, RL training, and data generation
& Infrastructure to develop environments and scale rollout collection, battle-tested in Nemotron 3
& Evaluation orchestration; benchmarks become judge agents assessing subject agents via A2A and MCP
& Framework for creating and sharing new RL environments; standardizes the environment side of post-training
& Evaluation and RL rollout framework; adapts existing benchmarks into a standard container-based format \\

Coverage
& 9 CUBEs (early stage); wraps any benchmark type including shared-infrastructure and VM-based ones
& 40+ environments across math, coding, tool use, and safety, and integrations with other environment libraries and benchmarks
& 250+ benchmarks across 17 domains, covering both static datasets (GAIA, SWE-Bench) and interactive environments
& 30+ environments across coding, games, web, and simulation; designed for new environment creation
& 46+ adapters for established benchmarks (SWE-Bench, Terminal-Bench, GPQA, ARC-AGI-2) with parity validation \\

Agent Interface
& MCP \texttt{tools/call} for actions; Gym-style \texttt{cube/reset}, \texttt{cube/step}, \texttt{cube/evaluate}
& Tools contract through OpenAI Responses API spec; flexible Agent scaffolding, easy to plug in any environment and Gym-like APIs implementable through HTTP servers
& A2A for task delegation; MCP for tool access; judge~+~subject~+~delegator roles
& HTTP Gym API (\texttt{reset}/\texttt{step}/\texttt{state}); MCP tool-calling interface for agent-environment interaction
& Wraps full coding agents (Claude Code, OpenHands, etc.); MCP servers configurable per task; ATIF trajectory format \\

RL Training
& Gym-compatible \texttt{evaluate}/\texttt{reset}/\texttt{step}; Python in-process for low-latency loops
& Ray-powered rollout; NeMo RL, OpenRLHF, Unsloth integrations
& Gym-to-MCP bridge available; primarily an evaluation platform
& TRL, TorchForge, Unsloth, SkyRL integrations; composable Rubric reward system
& QueueOrchestrator for dynamic RL loops; ATIF captures token IDs and logprobs \\

Adding a New Benchmark
& Implement a Python class once; works across all CUBE-compatible platforms via thin connectors
& Supports integrating benchmarks and creating new environments via separation of concerns between the Agent Harness and Environment Resources
& Implement a judge agent (A2A + MCP)
& Design a Docker environment with FastAPI and MCP tools; supports both new environments and wrappers around existing frameworks
& Write \texttt{task.toml} + \texttt{instruction.md} + Dockerfile + \texttt{test.sh}; Harbor-specific \\

Scale \& Deployment
& CUBE-harness uses Ray for parallel rollout; benchmarks declare resource requirements (\textit{what}), while pluggable backends dispatch to local or cloud infrastructure (\textit{how})
& Async first, server-based design; Ray for thousands of parallel rollouts
& Five operation modes from local development to hosted deployment; GitHub Actions CI used for competition leaderboards
& Docker locally; HuggingFace Spaces for sharing; Kubernetes for scaling
& Docker, Daytona, Modal, E2B, GKE; QueueOrchestrator for parallel trials \\

Registry \& Discovery
& Structured metadata: licenses, compliance badges, hardware requirements, runtime type
& Structured metadata: licenses, environment profiling, description, value; environments on GitHub, datasets on HuggingFace
& Registry for judge and subject agents; leaderboard and web UI
& HuggingFace Hub with \texttt{from\_hub()} discovery; tool registry planned
& \texttt{harborframework.com}; \texttt{dataset@version} versioning; 46+ curated adapters \\

\bottomrule
\end{tabular}
}
\end{table*}

The problem of benchmark fragmentation and the need for a unified evaluation infrastructure have been recognized by several communities, leading to a variety of platforms that attempt to address different aspects of this challenge \citep{bandel2026generalagentblog}. We organize our discussion by first examining domain-specific platforms that unify benchmarks within a particular task category, then surveying broader platforms that attempt cross-domain coverage, before presenting a detailed comparison of how these approaches relate to CUBE.

\subsection{Domain-Specific Unification Efforts}

Several platforms have emerged to standardize evaluation within specific task categories. BrowserGym \citep{ chezelles2025browsergym} provides a unified Gym-like environment for web agent research, integrating multiple web agent benchmarks under a common observation and action space. This ecosystem has significantly reduced fragmentation within the web agent community and demonstrated the value of standardized interfaces. However, BrowserGym is inherently limited to browser-based tasks and does not extend to other agent modalities such as terminal environments, desktop GUI control, or multi-modal reasoning tasks. Similarly, CUA-Bench \citep{cuabench2025} and related computer-use benchmarks focus specifically on desktop GUI interactions, while coding agent benchmarks like SWE-bench \citep{jimenez2023swebench} maintain their own evaluation harnesses. Each of these efforts represents valuable progress within its respective domains, but the proliferation of domain-specific standards compounds rather than resolves the overall fragmentation problem.
\subsection{Broader Platform Efforts}

Several platforms provide infrastructure for agent evaluation or training with some level of generality.

\paragraph{NeMo Gym} NVIDIA's NeMo Gym \citep{nemo-gym} provides a high-performance suite of RL training environments for domain-specific LLM tasks, including mathematics, science, coding, and tool-use. Rather than a specific environment interface, it introduces a highly scalable architecture that enforces a separation of concerns between the \texttt{Agent Scaffolding/Harness}, the \texttt{Environment Resources}, and the \texttt{LLM Model APIs}. By decoupling these components, NeMo Gym allows for independent scaling of compute-intensive models and complex environment simulations. NeMo Gym's contribution to standardization is mainly at the training harness layer, making it complementary to CUBE's focus on benchmark packaging and cross-platform portability.

\paragraph{AgentBeats} AgentBeats \citep{agentbeats2025} proposes an Agentified Agent Assessment (AAA) paradigm where benchmarks are realized as judge agents that evaluate subject agents through standardized A2A and MCP protocols, reducing N$\times$M agent-benchmark integrations to N+M protocol-level ones. The platform supports five operation modes from local development to hosted deployment, and hosts a public registry of assessments spanning coding, web, and multi-agent domains. These layers are composable: an AgentBeats judge agent could consume a CUBE-compliant benchmark through a thin connector, combining CUBE's portable infrastructure packaging with AgentBeats' evaluation protocol.

\paragraph{OpenEnv} Meta and Hugging Face's OpenEnv~\citep{openenv2025} provides a Gymnasium-style framework for creating and sharing RL training environments via a centralized HuggingFace Hub, with native integrations into TRL, TorchForge, Unsloth, SkyRL, and other training frameworks. An MCP tool-calling interface is available for agent-environment interaction, and a composable Rubric reward system is in active development. OpenEnv covers both new environments and wrappers around existing frameworks, including BrowserGym benchmarks (WebArena, VisualWebArena, WorkArena). Benchmarks requiring shared infrastructure must be provisioned externally by the user via environment variables; OpenEnv provides no lifecycle management for shared services or VM snapshots.

\paragraph{Harbor} Harbor \citep{Shaw_Harbor_Framework_2025} emerged from Terminal-Bench as a framework for evaluating agents and generating RL rollouts in container environments, with cloud deployment via Daytona, Modal, E2B, GKE, and Runloop. The framework integrates benchmarks through \emph{adapters} validated via parity experiments, and introduces the ATIF trajectory format capturing token IDs, logprobs, and tool definitions for RL and SFT pipelines; a QueueOrchestrator supports dynamic parallel rollout loops. MCP servers are configurable per task, and Harbor is designed to wrap full coding agents (Claude Code, OpenHands, Codex CLI, and others) rather than exposing a raw execution API. The per-task container model does not include lifecycle management for persistent shared infrastructure across tasks, which is reflected in the current adapter catalog not including benchmarks such as WebArena or OSWorld.

\paragraph{HAL} The Holistic Agent Leaderboard (HAL) \citep{kapoor2025holisticagentleaderboardmissing} from Princeton provides cost-controlled benchmarking across coding, web, science, and customer service domains. Its three-dimensional analysis (models, scaffolds, benchmarks) and LLM-aided log inspection offer valuable insights. However, it serves as an evaluation leaderboard rather than a training infrastructure, without Gym-compatible semantics, MCP-native tools, or support for RL rollout generation.

\paragraph{Exgentic} Exgentic, introduced in \emph{General Agent Evaluation}, is a practical framework for evaluating general agents across heterogeneous benchmarks through a Unified Protocol that mediates between agent interfaces and benchmark protocols. It is designed so that any supported agent can be run on any supported benchmark task while preserving native agent and benchmark behavior through external adaptors rather than intrusive modifications. The framework emphasizes scalable, reproducible evaluation, with support for parallel execution, isolated runs, standardized trajectories and cost reports, and an open general-agent leaderboard \citep{bandel2026generalagent, bandel2026agentic, bandel2026generalagentblog}. While Exgentic focuses on translating between heterogeneous existing protocols to enable unified evaluation, it does not primarily aim to specify the benchmark interface standard that should be adopted going forward; this is the layer CUBE targets.

\subsection{Comparison Summary}

Table \ref{tab:comparison} presents a detailed comparison of these platforms across key dimensions. The fundamental insight is that existing platforms have evolved from specific niches: NeMo Gym from RL training, AgentBeats from competition infrastructure, OpenEnv from the HuggingFace ecosystem, Harbor from SWE evaluation, and HAL from academic benchmarking. Each serves its origin community well, and together they address the agentic stack from complementary angles. The benchmark packaging and infrastructure lifecycle layer (how benchmarks declare resource requirements, manage shared services across tasks, and expose a portable interface independent of any particular harness) is the specific gap CUBE addresses by defining a minimal interface contract that any platform can implement, enabling benchmarks to be wrapped once and used everywhere, regardless of whether the downstream application is evaluation, training, or data generation.

\section{Alternative Views}

\paragraph{The Status Quo: Let Market Forces Decide}
The most common alternative maintains the current competitive landscape, letting natural selection determine dominance. However, platforms are already fragmenting by focus—evaluation vs. training, or domain-specific vs general—suggesting no single winner will emerge but rather multiple platforms adopted by research area. As noted in Section 1, without a shared interface, this competition skews towards integration catalog size, distracting from innovation while producing a divided ecosystem where platform choice determines benchmark access rather than technical merit. The ``wait for a winner'' strategy may simply result in permanent fragmentation organized by subfield rather than true consolidation.

\paragraph{Lighter-Weight Alternatives}
Rather than a comprehensive standard, the community could adopt lighter-weight solutions such as converter libraries that translate between existing platform formats or middleware layers that provide adapters without requiring changes to underlying benchmarks. Alternatively, the focus could shift from integration to curation, where the community selects a small canonical set of benchmarks that provide sufficient coverage of agent capabilities, eliminating the integration scaling problem by simply reducing the number of targets. This approach acknowledges that not all benchmarks need to be equally accessible and that research progress may be better served by depth on a few well-understood tasks rather than breadth across hundreds of environments. Yet converter libraries still require someone to write N-to-M translation layers, merely shifting the integration burden rather than eliminating it, and they introduce additional failure modes and maintenance overhead. Benchmark curation, while valuable, conflicts with the goal of building generalist agents that should succeed across diverse task distributions. Restricting evaluation to a small set risks overfitting our agent designs to those specific environments, and history suggests that canonical benchmark sets calcify and become divorced from real-world performance as the field advances.

\paragraph{Alternative Technical Designs}
CUBE's specific design choices are debatable. Critics might prefer explicit async primitives as first-class features, a simpler standard with fewer abstraction layers, pure Gym without MCP, or message-passing over RPC. The benchmark/task interface separation and centralized registry add complexity that may be unnecessary for some use cases. These concerns are valid. However, the key insight is not that CUBE's design is optimal, but that some standard is necessary. Building on established protocols like MCP and Gym minimizes the learning curve. The design should evolve through community feedback, but waiting for perfection ensures we never escape current fragmentation.

% --- SECTION 4: A CALL TO ACTION ---
\section{The Path Forward: A Call to Action}

The transition to a standard is never easy. It requires a collective agreement to prioritize interoperability over individual framework growth. However, the current trajectory of agent research is leading toward a fragmentation that will eventually stifle progress. We are spending too much time on DevOps and not enough time on AI.

We call on the authors of new evaluation platforms and benchmarks to join this effort. By adopting a standard like CUBE, we can ensure that every new benchmark is immediately available to the entire research community. We can create a world where a new breakthrough in agent architecture can be tested against hundreds of diverse environments within hours, rather than weeks.

The draft proposal presented here is just the beginning. We need a community-driven process to refine the API, define compliance levels, and build the registry. We invite researchers, developers, and platform owners to contribute to this discussion. The goal is shared infrastructure that lowers the barrier for everyone, not a mandate on how to do research.

\bibliography{references}
\bibliographystyle{icml2026}

\end{document}

%% file: commands.tex
\newcommand{\temp}[1]{\textcolor{gray}{[TBD] }}